\documentclass[sigconf, nonacm]{aamas} 

\usepackage{balance} 

\usepackage{amsthm}
\usepackage{amsmath}
\usepackage[ruled,vlined]{algorithm2e}
\usepackage{caption}
\usepackage{subcaption}
\usepackage{qtree}

\title[Designing Behavior Trees from Goal-Oriented LTLf Formulas]{Designing Behavior Trees from Goal-Oriented LTLf Formulas}
\titlenote{Accepted in Autonomous Robots and Multirobot Systems (ARMS) 2023 workshop affiliated with the 22nd International Conference on Autonomous Agents and Multiagent Systems (AAMAS 2023).}

\author{Aadesh Neupane}
\affiliation{
  \institution{Brigham Young University}
  \city{Provo}
  \country{USA}}
\email{adeshnpn@byu.edu}

\author{Eric G Mercer}
\affiliation{
  \institution{Brigham Young University}
  \city{Provo}
  \country{USA}}
\email{egm@cs.byu.edu}

\author{Michael A. Goodrich}
\affiliation{
  \institution{Brigham Young University}
  \city{Provo}
  \country{USA}}
\email{mike@cs.byu.edu}

\begin{abstract}
Temporal logic can be used to formally specify autonomous agent goals, but synthesizing planners that guarantee goal satisfaction can be computationally prohibitive. This paper shows how to turn goals specified using a subset of \textit{finite trace Linear Temporal Logic} (\LTL) into a \textit{behavior tree} (BT) that guarantees that successful traces satisfy the \LTL goal. Useful \LTL formulas for \textit{achievement goals} can be derived using achievement-oriented task mission grammars, leading to missions made up of tasks combined using LTL operators. Constructing BTs from \LTL formulas leads to a relaxed behavior synthesis problem in which a wide range of planners can implement the \textit{action} nodes in the BT. Importantly, any successful trace induced by the planners satisfies the corresponding \LTL formula. The usefulness of the approach is demonstrated in two ways: a) exploring the alignment between two planners and \LTL goals, and b) solving a sequential \textit{key-door} problem for a \textit{Fetch} robot. 

\end{abstract}


\newcommand{\BibTeX}{\rm B\kern-.05em{\sc i\kern-.025em b}\kern-.08em\TeX}

\newcommand{\LTL}{\textit{LTL}$_{\textit f}$ }

\newcommand{\PPATask}{\mbox{$\langle{\rm PPATask}\rangle$}}
\newcommand{\Mission}{\mbox{$\langle{\rm Mission}\rangle$}}

\newcommand{\LOne}{\mbox{$\langle{\rm L1}\rangle$}}
\newcommand{\LTwo}{\mbox{$\langle{\rm L2}\rangle$}}
\newcommand{\LThree}{\mbox{$\langle{\rm L3}\rangle$}}
\newcommand{\LFour}{\mbox{$\langle{\rm L4}\rangle$}}

\newcommand{\rectangle}{\fboxsep0pt\fbox{\rule{1em}{0pt}\rule{0pt}{1ex}}}

\begin{document}


\pagestyle{fancy}
\fancyhead{}


\maketitle 


\section{Introduction}
Specifying robot goals, creating plans, and verifying plans have received considerable attention in the research literature. Linear Temporal Logic (LTL) has been used in this context to specify system properties such as safety, liveness (something will keep happening), and goal-satisfaction~\cite{sistla1994safety}. However, the automata-based planning algorithms that accompany these systems often increase exponentially with an increase in the specification's length~\cite{pnueli1989synthesis}. Moreover, the tight coupling between LTL verification and planning problems with automata-based controllers leads to expensive re-computation even for small specification changes.  

One way to address these shortcomings is to decompose a complex goal specification into smaller modular specifications. 
Decomposing a complex goal specification using a Behavior Trees (BTs) is often useful because BTs are modular, maintainable, and reusable~\cite{colledanchise2018behavior}. 
Prior work demonstrated polynomial time correct-by-construction BTs from an LTL formula~\cite{colledanchise2017synthesis}, but a particularly restrictive specification format must be followed. The planning is also tightly coupled with the BT decomposition algorithm, making the integration of off-the-shelf planners impossible.  

This paper presents a mission-oriented grammar that generates \LTL formulas appropriate for sequential \textit{achievement goals}~\cite{van2008goals}. An algorithm is then presented that maps the \LTL formula into a corresponding BT, where the BT is structured to use Postcondition-Precondition-Action (PPA) structures. The PPA-style structure allows action nodes to be implemented with off-the-shelf planners. The paper demonstrates the usefulness of the resulting BTs using two case studies. First, the compatibility of a planner objectives with an \LTL goal is explored by designing plans using Markov Decision Process (MDP)-based planners and by constructing a simple sampling algorithm, and then comparing how well the outcome of the planners match the \LTL formula. Second, the \LTL-to-BT algorithm is used to construct \textit{Fetch} robot behaviors that successfully perform \textit{key-door} tasks when perturbations occur.



\section{Related Work}
LTL~\cite{pnueli1977temporal} is expressive enough to describe complex requirements (safety, liveness, goal-satisfaction) for discrete-time state transition systems. Bacchus et al.~\cite{bacchus1998planning} were the first to show that linear temporal logic not only can be used to specify system properties but also can be used to specify goals for formal logic systems. 

Verifying and synthesizing plans to satisfy complex goal specifications can be computationally prohibitive~\cite{pnueli1989synthesis,klein2006experiments}, but plans for restricted goal specifications can be found in polynomial time~\cite{piterman2006synthesis}. Thus, there appears to be a trade-off between the expressivity of the specification and the efficiency of the planners. One way to address this expressivity-computability trade-off is by decomposing expressive specifications into small modular pieces.
Decomposing complex plans into smaller pieces is not new. Most earlier methods applied decomposition not to the actual specifications but to supporting algorithms ranging from parsing to planning, including reinforcement learning-based approaches~\cite{barnat2002distribute,rozier2011multi,maretic2014ltl,sutton1999between,parr1998reinforcement}. Colledanchise et al.~\cite{colledanchise2017synthesis} were the first to demonstrate direct decomposition of LTL specifications. Vazquez et al.~\cite{vazquez2018learning} went in a reverse direction  by demonstrating a method to construct temporal LTL specifications from a grammar containing atomic specifications. 

Frequently, decomposing a LTL specification and creating planners are done concurrently. Generally, the planning process involves: a) converting LTL goal specification to an automaton, b) creating an automaton modeling the environment, c) constructing a product automaton, d) playing the Rabin game using game theory concepts, and e) discretizing the plans~\cite{antoniotti1995discrete,fainekos2005temporal,bertoli2001mbp,jensen2000obdd, fainekos2005hybrid,toro2018teaching,schillinger2018decomposition,tadewos2022specification}. Probabilistic Computation Tree Logic is an alternative sometimes used when the planning process is computationally expensive and when uncertainties are present~\cite{lahijanian2010motion,ding2011ltl}. When the guarantees provided by automaton-based planning are not required, sampling-based motion planning algorithms can be used~ \cite{vasile2013sampling,ahmadi2020stochastic}.  Interestingly, the acceptance of the trace by the automaton can be used as a reward function in some MDP problems~\cite{sadigh2014learning}. 

BT representations are (i)~equivalent to Control Hybrid Dynamical Systems and Hierarchical Finite State Machines~\cite{marzinotto2014towards} and (ii)~generalizations to the Teleo-Reactive paradigm and And-Or-Trees~\cite{colledanchise2016behavior}. BT modularity can be combined with the verification properties of LTL. Biggar et al.~\cite{biggar2020framework,biggar2022modularity} developed a mathematical framework in LTL is used to verify BT correctness without compromising valuable BT traits: modularity, flexibility, and reusability. 

\section{Behavior Tree and \LTL Essentials}
Assume a standard state transition system $\mathbb{T}: S\rightarrow S$, where $\mathbf{s}_t= [s_1, s_2, \dots, s_n]\in S$ is a vector of state variables at time~$t$. A \textit{trace} is a sequence of states $\tau=\langle\mathbf{s}_0,\mathbf{s}_1, \dots,\mathbf{s}_t\rangle$. 
Trace segments are indexed using the following notation: $\tau[i] = \mathbf{s}_{i}$ and $\tau[i:j] = \langle\mathbf{s}_i, \dots,\mathbf{s}_{j}\rangle$. \LTL works on finite traces so all traces should be less than or equal to the \textit{max-trace-length} $m$. i.e., $|\tau| \leq m$.


\subsection{\textbf{Behavior Tree Overview}}
\begin{table}[tb]
\centering
 \begin{footnotesize}
 \begin{tabular}{|l|c|c|l|l|}
 \hline
 {\bf Node} & Figure & Symbol& \footnotesize{\bf Success} & \footnotesize{\bf Failure} \\
 \hline\hline
  \textbf{Sequence} & \textbf{$\rightarrow$} & $\sigma$ & {all children succeed} & {one child fails} \\ 
 \hline
 \textbf{Parallel} & \textbf{$\rightrightarrows$} & $\pi$ & {all children succeed} & {one child fails} \\ 
 \hline
 \textbf{Selector} & \textbf{$?$} & $\lambda$ & {one child succeeds} & {all children fail} \\ 
  \hline
 \textbf{Decorator} & $\Diamond$ & $\delta$  &\multicolumn{2}{|c|}{User defined}\\ 
 \hline
 \textbf{Action} & $\rectangle$ & $\alpha$ & task complete & task failed\\ 
 \hline
 \textbf{Condition} & $\bigcirc$ & $\kappa$ &  true & false \\ 
  \hline
\end{tabular}
\end{footnotesize}
  \caption{ BT node type notation. }
  \label{tab:BT_nodes}
\end{table}

A BT is a directed rooted tree where the internal nodes are called \textit{control nodes}, and the leaf nodes are called \textit{execution nodes}. A node can be in only one of three states: \textit{running} (processing is ongoing), \textit{success} (the node has achieved its objective), or \textit{failure} (anything else). 
A \textit{blackboard} memory holds relevant data that can be passed between BT nodes~\cite{shoulson2011parameterizing}. 
This paper uses a simple hash-table blackboard with a global scope. BT execution is governed by the root node, which generates \textit{signals} at a particular frequency. Signals are recursively sent from a control node to its children. In this paper, BT execution halts if the root node returns either success or failure. 

This paper uses the formulation in~\cite{colledanchise2018behavior}, which has two types of execution nodes, \textit{Action} and \textit{Condition}.  Action nodes execute actions of plans in the world, and condition nodes evaluate some property of the world (e.g., check proposition). 
There are four control node type: \textit{Sequence}, \textit{Selector}, \textit{Parallel}, and \textit{Decorator}.  
Table~\ref{tab:BT_nodes} summarizes the graphical symbols, mathematical notation, and behaviors of all node types. By treating success as true and failure as false, \textit{selector} node types act as a logical $\vee$ and both \textit{sequence} and \textit{parallel} node types act as a logical $\wedge$,  except for when one of its inputs is the \textit{running} state. Sequence nodes run the leftmost child until it succeeds/fails before subsequent nodes, in contrast to parallel nodes which run all nodes. The decorator node $\delta$ has a single child that manipulates the return status of its child according to a user-defined rule. All node types return \textit{running} for any condition not in the table. In this paper, the condition nodes represent propositions, so they can only return success or failure. Only actions and control nodes can be in all three states.

\subsubsection*{\textbf{Semantics}}
\label{sec:bt_semantic}
Let \textit{N}$\in\{\sigma,\pi,\lambda,\delta,\kappa\}$ denote either a condition node type or one of the control node types from Table~\ref{tab:BT_nodes}. Let $\sigma N_1, N_2$ denote a sequence sub-tree,  $\lambda N_1, N_2$ denote a selector sub-tree, $\pi N_1, N_2$ denote a parallel sub-tree and $\delta N$ denote a decorator sub-tree. Note, for control nodes $\pi, \lambda, \sigma$ their child nodes $N_1,N_2,\dots, N_n$ are ordered from left to right as the control flows in BT from left to right. The input to the BT is the environment state $\mathbf{s}_t$. A BT condition node is an encoding of the state $\mathbf{s}$ in predicate logic. Let \textit{A} be proposition variables which is the result of the predicate logic evaluated on $\mathbf{s}$. BT node \textit{N} that returns success is denoted by $\models \mathbf{s} \; N$ and is inductively defined in terms of other BT nodes. Recall, in a BT, the leaf nodes are always the condition or action nodes.

\begin{tabular}{|l|l|l|}\hline
     \textbf{BT Node} &  \textbf{Definition} & \textbf{Note} \\ \hline\hline
$\models s \ \kappa$ &  $\models s \ \phi_{\kappa}$ & $\phi_\kappa$ \mbox{ is condition expressed} \\ && \mbox{in predicate logic} \\ \hline
$\models s \ \delta N$ &  $\models s\ f{\delta}(N)$ & $f{\delta}$ \mbox{ is function that operates} \\ && \mbox{on $N$'s return status}\\  \hline

$\models s \ \delta{\neg} N$ & $\not\models s\ N$ & \mbox{negation decorator node}\\ \hline
$\models s \ (\pi \ N_{\rm L} \ N_{\rm R})$  &  $\left( \models s\ N_{\rm L}\right)\ \wedge$ \\ &  $\left( \models s\ N_{\rm R}\right)$ & $\wedge$ \mbox{ boolean {\tt AND} operator}\\ \hline
$\models s \ (\sigma\ N_{\rm L} \ N_{\rm R})$ &  $\left( \models s\ N_{\rm L}\right)\ \wedge^*$ \\ & $\left( \models s\ N_{\rm R}\right)$ & $\wedge^*$ \mbox{ boolean {\tt AND}, but $\left( \models s\ N_{\rm R}\right)$} \\
 && \mbox{\ \ \ \ not evaluated  when $\left( \not\models s\ N_{\rm L}\right)$}\\ \hline
$\models s \ (\lambda\ N_{\rm L} \ N_{\rm R})$ &   $\left( \models s\ N_{\rm L}\right)\ \vee^*$ \\ & $\left( \models s\ N_{\rm R}\right)$ & $\vee^*$ \mbox{ boolean {\tt OR}, but $\left( \models s\ N_{\rm R}\right)$} \\
 && \mbox{\ \ \ \ not evaluated  when $\left(\models s\ N_{\rm L}\right)$}\\ \hline
\label{tab:bt_semantics}
\end{tabular}

\subsection{\textbf{\LTL Overview.}}
\label{sec:ltlf}
\LTL uses a finite set of {propositional variables}; the logical operators $\neg$, $\vee$ , and $\wedge$; and the temporal modal operators \textbf{X} (``neXt''), \textbf{U} (``Until"), \textbf{F} (``Finally", meaning at some time in the future), and \textbf{G} (``Globally"). 

\subsubsection*{\textbf{Syntax}}
In \textit{LTL}$_{\textit{f}}$, unary operators take precedence over binary, the \textbf{U}ntil operator takes precedence over the $\wedge$ and $\vee$ operators, and all operators are left associative. Each state variable $s_i$ in the state vector $\mathbf{s}_t$ represents an assignment of the $i^{\rm th}$ atomic proposition to {\tt true/false}. An \LTL formula is formed from a set of $n$ atomic propositions $A\in P_{\rm atomic}$ by the following grammar, which is unambiguous, enforces precedence, and enforces left associativity:
\begin{eqnarray*}
    \psi &:=& \LOne\  | \ \vee \psi \  \LOne \\
    \LOne &::=& \LTwo \ | \ \wedge \LOne \ \LTwo \\
    \LTwo &:=& \LThree \ | \ \textbf{U} \LTwo \ \LThree \\
    \LThree &::=& \LFour \ |\ \neg\LThree\ |\ \textbf{X}\LThree\ |\ \textbf{F}\LThree\ |\ \textbf{G}\LThree \\
    \LFour &::=& A\ |\ (\psi)
    \label{eq:Unambiguous_grammar}
\end{eqnarray*}

\subsubsection*{\textbf{Semantics}}
The semantic interpretation of an \LTL formula, $\psi$, is given using the \textit{satisfaction relation}, $\models$, which defines when a trace satisfies the formula. Recall that $\tau[i]=\mathbf{s}_i$ is a vector of state variables. Thus a trace is a sequence of boolean state variable values. Let $m$ denote the maximum length of a finite \LTL formula. Thus, a full trace is $\tau[0:m]$ and the ``suffix'' of a trace beginning at time $i$ in $0<i\leq m$ is $\tau[i:m]$. A trace suffix $\tau[i:m]$ satisfying an \LTL formula $\psi$ is denoted using prefix notation by $\models \tau, i \ \psi$ and is inductively defined (using the $\equiv$ to indicate ``defined'') as follows:  
\begin{itemize}
\setlength\itemsep{-.08cm}
    \item $\models \tau, i \; A \equiv A \in \tau[i]$ and $A\equiv{\tt true}$
    \item $\models \tau, i \; (\neg \psi) \equiv (\not \models \tau,i \; \psi)$
    \item $\models \tau,i \; (\wedge \; \psi_1 \; \psi_2) \equiv \ \models  \tau, i \; \psi_1$ and $\models \tau,i \; \psi_2$.
    \item $\models \tau, i \; (\vee \; \psi_1 \; \psi_2) \equiv \ \models  \tau,i \; \psi_1$ or $\models \tau, i \; \psi_2$.
    \item $ \models \tau, i  \; (\textbf{X} \; \psi) \equiv \ \models \tau, (i+1) \; \psi$.
    \item $\models \tau, i \; (\psi_1 \; \textbf{U} \; \psi_2) \equiv \ \models \; \tau,k \; \psi_1$ and $\models \; \tau,j \; \psi_2$ where $\exists j\in\{i,i+1, \ldots, m\}$ such that $\forall k\in \{i,i+1,\ldots,j-1\}$.
    \item $\models \tau,i \; (\textbf{G} \; \psi) \equiv \ \models \; \tau,k  \; \psi$ where $\forall k\in[i,m]$.
    \item $\models \tau,i \; (\textbf{F} \; \psi) \equiv \ \models \;\tau,k \; \psi$ where $\exists k \in[i,m]$.
\end{itemize}

\section{\LTL Task Grammar and Resulting BTs}
This paper restricts attention to goal specifications that use a special \textit{task} and \textit{mission} grammar which follow a postcondition, precondition, action (PPA) structure.  PPA structures have proven useful for agent-based and robotics applications. The resulting subset of \LTL that uses the PPA structure is called \textit{PPA-LTLf}.
This section shows that a BT can be constructed from the \textit{PPA-LTLf} formula that \textit{generates} traces that provably satisfy the \textit{PPA-LTLf} formula under certain conditions. The proof is outlined below and applies to (a)~missions that sequence tasks using \LTL operators and (b)~tasks that use the PPA structure from the BT literature, which checks the postcondition before trying an action~\cite{colledanchise2018behavior}.

\subsubsection*{\textbf{PPA-Style Task Formula}}
\label{sec:task_grammar}
A general achievement goal in the PPA structure can be represented using a simple proposition logic formula $PPATask = (PoC) \vee (PrC \wedge Action)$ where \textit{PoC}, \textit{PrC}, \textit{action} are postcondition, precondition, and action propositions. The $\vee$ operator ensures that the action is not executed if the postcondition is already satisfied. If the postcondition is not satisfied, then the $\wedge$ operator ensures that action is only executed when the precondition is satisfied. Drawing from real-world applications, the goal specifications can be generalized using the \LTL formula in equation~\ref{eq:PPATask}. 

\begin{eqnarray}
    \psi^{PPATask} &:==& \vee [\wedge (\textbf{G} \ \it{GC}) \; \it PoC](\wedge (\wedge (\textbf{G} \ \it{GC}) \; (\textbf{F} \ \it{PrC})) \nonumber \\ 
    && (\textbf{U}(\it{TC}) \; (\wedge \; (\textit{Action}) \; (\textbf{G} \ \it{HC}))))
    \label{eq:PPATask}
\end{eqnarray}
where the notation PoC, PrC, GC, TC represent any postcondition, precondition, global, or task constraints propositions, respectively. We'll elaborate on what it means to say that {\tt Action}$=${\tt true} in a subsequent section, but for now we simply state that {\tt Action}$=${\tt True} if and only if the postcondition is satisfied.

\newcounter{tmp}

\subsubsection*{\textbf{BT for a PPA-Style Tasks}}
\begin{figure}[htb]
  \centering %
	\includegraphics[width=.99\linewidth, keepaspectratio]
	{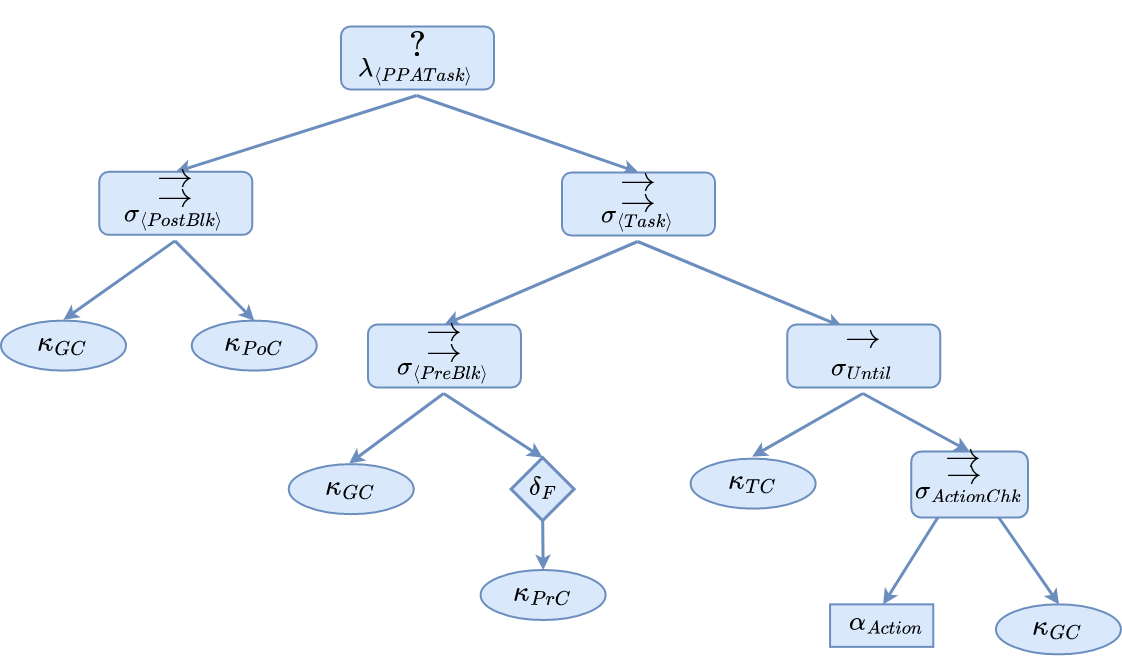}
    \caption{\small General BT $\mathbb{T}_{\langle{\rm PPATask}\rangle}$ for any PPATask $\psi$.}
    \label{fig:generalBTPPATask}
\end{figure}

Let $\psi$ denote a PPATask formula and let $L(\psi)$ denote the set of all traces that satisfy~$\psi$. Figure~\ref{fig:generalBTPPATask} shows the \textit{task behavior tree}, denoted by $\mathbb{T}_\psi$, for the task formula~$\psi$. The BT $\mathbb{T}_\psi$ executes a plan to generate traces~$\tau$ that satisfy the task formula~$\psi$. The $L(\mathbb{T}_\psi)$ denote the set of all trace that can be generated by $\mathbb{T}_\psi$ when $\mathbb{T}_\psi$ return success. The structure and semantics of  $\mathbb{T}_\psi$ allows only a subset of traces that satisfy~$\psi$ to be generated, i.e., $L(\mathbb{T}_\psi) \subseteq L(\psi)$.

The structure of the tree in Figure~\ref{fig:generalBTPPATask} follows directly from the task formual in equation~\ref{eq:PPATask}. Each binary operator is represented as a subtree that is rooted at either a sequence node or a selector node, and each of these subtrees has two branches, one for each of the two operands.  Each constraint or condition is represented by a condition node and, except for the precondition, which has a decorator node as a parent. Each subtree is now described.


\textbf{Subtrees for $\wedge$ and $\vee$ Operators}. Each $\wedge$ operator is implemented using a parallel node ($\rightrightarrows$), and the $\vee$ operator is implemented using a selector node (?). The root selector node indicates that a successful trace is generated either because the postcondition is satisfied in the world (left branch) or because the agent takes an action that satisfies the postcondition.

\textbf{Subtrees for Postconditions and Preconditions.} The postcondition is represented only by a condition node (oval) because anytime the postcondition is satisfied without violating the global constraint then the trace is successful. The precondition is represented by a condition node with a decorator ($\delta_{F}$) node parent (diamond) whose operation remembers if the precondition was satisfied when task execution began.

\textbf{Subtree for Global Constraints.}
Each constraint is represented by a condition node that checks the boolean formula that encodes the constraint. Global constraints must be satisfied for each stage of execution, so the global constraint block is repeated when checking the postcondition, the precondition, or the task execution. Once the global constraint is not satisfied, it returns failure, then that return signal propagates to the root node. Recall that when the root node is either in a success or failure state, the BT execution is halted.

\textbf{Subtree for Actions.}
An action is encoded in an action node type (rectangle). A time limit $T_{\rm TaskMax}$ is imposed on plan execution since only finite-length traces are permitted. The $\alpha_{Action}$ node can return success, failure or running based on the following rules:
\[
  \alpha_{Action}[t] = 
  \left\{
    \begin{array}{ll} 
      \mbox{\textit{success}} & \mbox{ if $({\rm PostCon} = {\tt true})$ and $ (t \leq T_{\rm TaskMax}) $} \\
      \mbox{\textit{running}} & \mbox{ if $({\rm PostCon} = {\tt false})$ and $(t < T_{\rm TaskMax}) $} \\
      \mbox{ \textit{failure}} & \mbox{otherwise}
    \end{array}
  \right.
\]

\textbf{Subtree for the Until Operator.}
The \textbf{U}ntil operator is implemented using a sequence node type, $\sigma_{\it Until}$, with two children subtrees. The \textit{right} child subtree of $\sigma_{\it Until}$ is a subtree rooted at a sequence node that executes the action and evaluates whether the global constraint is violated.
The \textit{left} child is the task constraint, which consists of condition node~$\kappa_{{\rm TCnstr}}$. The task constraint must be satisfied for all time up to when the plan node, $\alpha_{\it Action}$, is succeeds, but the task constraint need not be satisfied once the postcondition is satisfied.

\section{BT for \LTL Task-based Missions}
\label{sec:missionBT}
A mission is defined as a set of tasks that are organized using the LTL operators. Therefore, the BT for a mission includes subtrees composed of the BTs for PPA-style tasks. Combining PPA-style tasks using the LTL operators is constrained by both (a)~the nature of tasks and (b)~the task BT implementation of a PPA-style task. This paper constrains what a mission can accomplish. First, the Ne\textbf{X}t operator is not included in the mission grammar because the variability of task execution can prevent precise timing between tasks. Instead, sequencing tasks in particular orders is implemented using \textbf{F}inally and \textbf{U}ntil.

Second, an external perturbation might ``undo'' the postcondition previously accomplished by the task plan. It is therefore desirable to rerun the plan provided that the precondition is satisfied when the plan begins to run a second time. Unfortunately, the PPA BT ${\mathbb T}_\psi$ remembers whether the precondition was satisfied when the task first began. 
This means that subsequent calls to the task do not have the ability to check the precondition, and an important check is lost. Fortunately, if the plan can accomplish the task without checking the precondition then the mission can still succeed.

Third, it is possible for two PPA-style tasks to compete for the same resources (e.g., try to use the same actuator at the same time).  Both resource allocation issues and very precise timing between tasks must be solved by a single planner that coordinates both tasks. 

\subsubsection*{\textbf{Mission Grammar}}
The grammar expressed in Productions~(\ref{Production:M1})--(\ref{Production:M8})) uses prefix operator notation. Productions~(\ref{Production:M1}), (\ref{Production:M3}), and (\ref{Production:M5}) enforce precedence: \textbf{F}inally has higher precedence than \textbf{U}ntil, which has higher precedence than $\wedge$, which has higher precedence than~$\vee$. Productions~(\ref{Production:M2}), (\ref{Production:M4}), and (\ref{Production:M6}) enforce the left associativity required of the binary LTL operators. Production~(\ref{Production:M8}) means that a Mission is composed of one or more PPA-style tasks. The parentheses in Productions~(\ref{Production:M7})--(\ref{Production:M8}) ensure that tasks in a multi-task mission are distinct.
\setcounter{tmp}{\value{equation}}
\setcounter{equation}{0}
\renewcommand{\theequation}{M\arabic{equation}}
\begin{eqnarray}
\label{missiongrammar}
    \Mission & ::=& \LOne \ \label{Production:M1} \\
    & ::=& \vee \Mission \LOne \label{Production:M2}\\
    \LOne & ::=& \LTwo \  \label{Production:M3}\\
     & ::=& \wedge \LOne  \LTwo \label{Production:M4}\\
    \LTwo & ::=& \LThree \  \label{Production:M5}\\
     & ::=& {\textbf U} \LTwo  \langle {\rm  L3} \rangle\label{Production:M6}\\
     {\LThree} & ::=& {\textbf F} \ \Big( \Mission \Big) \label{Production:M7}\\
     & ::=& \Big(\PPATask\Big)  \label{Production:M8}
\end{eqnarray}
\renewcommand{\theequation}{\arabic{equation}}
\setcounter{equation}{\value{tmp}}

\subsubsection*{\textbf{Constructing the Mission BT}}
Let $\Psi$ (capital $\psi$) denote a mission formula derived from the grammar above and expressed in \LTL form. A parser for $\Psi$ can be used to produce the expression tree for the formula, where each internal tree node is an \LTL operator from the mission grammar and each leaf node is a PPA task subtree, ${\mathbb T}_\psi$ referred to by Production~(\ref{production:PPATask}). The mission BT ${\mathbb T}_{\Psi}$ is constructed from the expression tree in the following manner.

\textbf{Root node of ${\mathbb T}_{\Psi}$.} The root of the tree is a decorator node denoted by $\delta_{\langle{\rm Mission}\rangle}$. This decorator node  not only passes through the status of its child node but also tracks time, returning \textit{failure} if $T_{\rm TaskMax}$ is reached. Thus, running out of time is detected by either the $\delta_{\langle{\rm Mission}\rangle}$ node in the mission BT or the action node $\alpha_{\textit{Action}}$ node in the task BT.

\textbf{Leaf nodes of${\mathbb T}_{\Psi}$.} Each leaf node is represented by a decorator node $\delta_{\langle{\rm PPATask}\rangle}$ that passes through the value returned by its child. This decorator node separates each task BT, ${\mathbb T}_{\psi}$, from the rest of the tree, effectually enforcing the parentheses in Production~(\ref{Production:M8}).

\textbf{$\boldsymbol{\vee}$ Operator}. Production~(\ref{Production:M2}) says that some missions can be accomplished in multiple ways, \textit{either} by performing the combination of tasks in the tree descending from the left branch \textit{or} the right branch. The right-hand side of Production~(\ref{Production:M2}) is $\vee \Mission \LOne$, which is implemented as a subtree rooted at a selector node, $\lambda_{\langle{\rm Mission}\rangle}$. 



\textbf{$\boldsymbol{\wedge}$ Operator}. Production~(\ref{Production:M4}) says that some missions require multiple tasks to be performed. The right-hand side of the production is $\wedge \LOne \LTwo$, which is implemented as a subtree rooted at a parallel node type, $\pi_{\LOne}$. The parallel node allows each subtree structure to check to see if success occurs each time step.  

\begin{figure}[htb]
  \centering %
	\includegraphics[width=.5\linewidth, keepaspectratio]{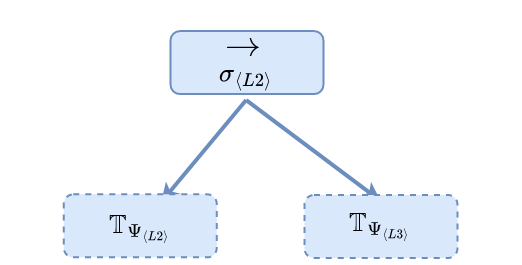}
    \caption{\small Mission BT $\mathbb{T}_{\Psi}$ for ${\mathbf U}\ \psi_{\LTwo} \  \psi_{\LThree}$. } 
    \label{fig:PsiPPATaskUntil}
\end{figure}

\textbf{Until Operator}.
Production~(\ref{Production:M6}) says that some missions require (a)~the task determined by applying productions rooted at $\LTwo$ to succeed performed (b)~\textit{until} the task determined by applying productions rooted at $\LThree$ succeeds. Figure~\ref{fig:PsiPPATaskUntil} shows the Until sub-tree, which is rooted in a sequence node $\sigma{\LTwo}$ consistent with the Until sub-tree for PPATask BT $\mathbb{T}_{\psi}$. Recall the semantic of sequence node from the table~\ref{tab:bt_semantics}, the right leaf node is only executed if the left node is successful, which matches to the  \LTL semantics for \textit{Until} operator described in section~\ref{sec:ltlf}.


\textbf{Finally Operator}. Production~(\ref{Production:M8}) uses the finally operator. The semantics of the finally operator in mission context are that if the child subtree, $T_{\langle{\rm Mission}\rangle}$, fails then it should be given another chance to succeed. 

The finally operator is implemented using a decorator node, $\delta_{\textbf{F}}$, that returns success as soon as its child subtree ${\mathbb T}_{\langle{\rm Mission}\rangle}$ returns success or if the child has ever previously returned success. If the child subtree ${\mathbb T}_{\langle{\rm Mission}\rangle}$ has not previously succeeded and returns running, then $\delta_{\textbf{F}}$ returns running. If the child subtree ${\mathbb T}_{\langle{\rm Mission}\rangle}$ returns failure, then the decorator node returns running and also resets any \textit{descendant} decorator nodes. Two significant differences exist between the $\delta_{\textbf{F}}$ decorator nodes in mission and task BT. The first difference is the child node of the $\delta_{\textbf{F}}$ decorator is always a condition node for task BT, whereas $\mathbb{T}_{PPATask}$ is the child node for mission BT. The second difference is that the $\delta_{\textbf{F}}$ decorator node has the reset method, which the task BT lacks. The $\delta_{\textbf{F}}$ has memory to track how often the reset signal has been passed to its child node. $\delta_{\textbf{F}}$ can only send \textit{max-reset-threshold} $(\theta)$ times reset signals so that the BT is bounded and does not run forever. 
Note that the $\delta_{\textbf{F}}$ decorator node separates out the subtree descending from the $\Mission$ nonterminal from the rest of the BT, effectually enforcing the parentheses in Production~(\ref{Production:M7}).


In summary, there are two outputs from the finally decorator node: the status returned to its parent and the \textit{reset} command to its descendent PPA BTs. The reset command is only issued if ${\mathbb T}_{\langle{\rm Mission}\rangle}[t] = \textit{failure}$.  The status behavior is defined as 
\[
    \delta_{\textbf{F}}[t] = \left\{
        \begin{array}{ll}
            \mbox{\textit{success}} & \mbox{if } {\mathbb T}_{\langle{\rm Mission}\rangle}[t] = \mbox{\textit{success}} \\ 
            \mbox{\textit{failure}} & \mbox{if ($t \geq T_{\rm \theta})$}\\
            \mbox{\textit{running}} & \mbox{otherwise}
        \end{array}
    \right.
\]



\subsubsection*{\textbf{Proof Outline}}
There is not enough room to show the complete proof that every successful trace generated by ${\mathbb T}_{\Psi}$ satisfies the \LTL formula $\Psi$ if $\Psi$ is formed by the mission and task grammars\footnote{For complete proof refer the authors dissertation~\cite[Chapter~5]{neupane2023designing}}. The proof outline is easy to follow. 

\textbf{Successful Task Traces.} Each task formula $\psi$ produced by the task grammar must satisfy Eq.~(\ref{eq:PPATask}). This equation has the outcome of an action node type plus postconditions, preconditions, global constraints, and task constraints. An abstract trace can be constructed of the truth values of the conditions and constraints and the return values of the action nodes. It is straightforward exercise to show that every successful abstract trace generated by the task behavior tree ${\mathbb T}_{\psi}$ depends on the correct truth values of the constraints and conditions happening at the appropriate time to satisfy the \LTL task formula given in Eq.~(\ref{eq:PPATask}).

\textbf{Successful Mission Traces.} The proof that every successful mission trace satisfies a mission formula~$\Psi$ is inductive. The base case is for a mission consisting of a single task, which reduces to the proof that each successful trace from the task BT ${\mathbb T}_{\psi}$ satisfies the task formula~$\psi$. There is then an inductive step for the $\vee$ operator and the $\wedge$ operator, which is straightforward because these operators map naturally to selector and sequence nodes. There is an inductive step for the \textbf{U}ntil operator, which is a bit more complicated because of the timing of when the two tasks succeed but which can be established with careful bookkeeping of successful traces. The last inductive step is for the \textbf{F}inally operator, which is straightforward since the operator is implented as an easy to understand decorator node.

\section{Planner-Goal Alignment Example}
\label{sec:planner-goal-misalignment}

Given the BT produced from a given mission formula, it is necessary to choose a plan or policy for each action node. Traditional LTL plan synthesis using advanced automata (Rabin and Buchi) requires a suitable environment model and is of high computation complexity. When the goal specifications are valid but sufficient information about the environment is unknown, traditional automata-based solutions are infeasible. However, the BT does not specify what type of planner is required. Instead, off-the-shelf planners can be used to design the plans or policies used in the BT's action nodes.  This is demonstrated using two planners: policy iteration, which uses reward functions to represent goals that may or may not align with the the \LTL formula, and a state-action table that uses the BT return status to update the probability of actions in successful or unsuccessful action sequences. 



\subsection{Problem Formulation}
The \textit{Mouse and Cheese} problem is a classic grid world problem~~\cite{russell2010artificial}. This paper uses a 4x4 grid giving sixteen world locations indexed by $s_{j,k}$. There is one atomic proposition $A_{j,k}$ for each state, where $A_{j,k}=$\textit{true} indicates that the mouse occupies location~$s_{j,k}$. The mouse may not occupy two locations simultaneously, $A_{i,j} == {\it true} \rightarrow A_{k,\ell} == {\it false}$. Location~$s_{4,4}$ contains the cheese, and the mouse automatically picks up the cheese when it occupies that cell. An atomic proposition $Cheese=$\textit{true} indicates that the mouse has the cheese. Location~$s_{4,2}$ is dangerous, denoted by atomic proposition $Fire$, and the mouse should avoid this state. Atomic proposition $Home$ indicates whether the agent is at home location~$s_{3,1}$. The state vector at time~$t$ is the vector of truth values for each atomic proposition $\mathbf{s}_t = [s_{1,1},s_{1,2},\ldots, s_{4,4},\mbox{\it Cheese}, \mbox{\it Fire}, \mbox{\it Home}]$.

Paraphrasing from~\cite{russell2010artificial}, the agent has four actions: move up, left, right, or down. if the agent bumps into a wall, it stays in the same square. The agent's actions are unreliable, i.e., the `'intended'' action occurs with some probability $p_{\rm in}$ but with some lower probability, agents move at the right angles to the intended direction, $1-p_{\rm in}$. 
The problem gets harder for the agent as $p_{\rm in}$ decreases.  

The mouse and cheese problem is a sequential planning problem, requiring the agent to find the cheese and then return it to the home location. This is an achievement goal that requires two separate plans (or policies): (a)~find the cheese avoiding fire and (b)~return home avoiding fire after finding the cheese. There are many planners that can solve this problem (e.g., MAX-Q and other hierarchical learners), and the point of this section is not to argue for the best way to solve the problem. Rather, the point is to explore how well different types of planners, one for each task, align with the overall mission goal.  

The goal of the mouse is to retrieve the cheese while avoiding the fire location and getting back at the home location. An \LTL formula from the mission grammar for the sequential find-the-cheese-and-return home (C2H) and the corresponding PPA task formulas from Eq.~(\ref{eq:PPATask}) are
\begin{eqnarray*}
    \Psi^{\rm C2H} &=& \textbf{U }\textbf{F} \psi^{\rm Cheese} \;  \textbf{F} \psi^{\rm Home}  \\
    \psi^{\rm Cheese} &=& \vee (\wedge \neg\it{Fire} \ \it{Cheese})(\wedge (\wedge \neg{\it Fire} \; {\it True}) \\
    && (\textbf{U} \; {\it True} \; \wedge \; \textit{Action}_{\rm Cheese} \; \neg\it{Fire}))\\
    \psi^{\rm Home} &=& \vee (\wedge \neg\it{Fire} \ {\it Home})(\wedge (\wedge \neg\it{NotFire} \; {\it Cheese}) \\
    && (\textbf{U} \; {\it True} \; \wedge \; \textit{Action}_{\rm Home}))
\end{eqnarray*}
The BT action nodes $\textit{Action}_{\rm Cheese}$ and $\textit{Action}_{\rm Home}$ execute the plans that lead mouse to the cheese and return home, respectively.

\subsection{Action-Node Policies via Policy Iteration}
This section uses policy iteration to create a policy for ${\it Action}_{\rm Cheese}$ and again to create a policy for ${\it Action}_{\rm Home}$. The reward structure for the cheese task is ${\mathbf r} = (r_{\rm other},r_{\rm cheese},r_{\rm fire})$, where the rewards are for for occupying any grid cell other than home or fire, having the cheese, and occupying the fire cell, respectively. The reward structure for the home task is ${\mathbf r} = (r_{\rm other},r_{\rm home},r_{\rm fire})$, where the rewards are for for occupying any grid cell other than home or fire, occupying the home cell, and occupying the fire cell, respectively. Restrict attention to situations where $r_{\rm cheese}=r_{\rm home} = r_{\rm good}$, which allows results to be represented using a triple  ${\mathbf r} = (r_{\rm other},r_{\rm good},r_{\rm fire})$.

Using policy iteration to create policies from rewards encodes goal in two ways: once in the BTs via the return values of the root node, and once in the reward structures themselves. The reward-goal alignment problem is well-known, and the problem is explicit when the goal is encoded directly in the \LTL formula but indirectly in the reward structures. By construction of the BT, every successful trace satisfies the goal, but many traces fail or time-out depending on the reward structure. 

To illustrate, five hundred twelve independent experiments are conducted for various intended action probabilities $p_{\rm in}$ and rewards values. The dependent variables are \textit{success probability}, which is defined as the number of simulations where $\Psi^{C2H}$ is satisfied divided by the total number of simulations, and \textit{trace length}, which is the length of the trace. Experiments used the following:
\begin{center}
    \begin{tabular}{c|c}
        \textbf{Parameter} & \textbf{Values} \\\hline
        $r_{\rm other}$ & $\{-1.5, -1.4, -1.3, \ldots, -0.2, -0.1, -0.04\}$ \\
        $r_{\rm cheese} = r_{\rm home}$ & $\{0.1,0.5, 1.0, 2.0, 5.0, 10.0 \}$ \\
        $r_{\rm fire}$ & $\{-10, -5, -2, -1, -0.5, -0.1 \}$\\
        $p_{\rm in}$ & $\{0.4, 0.45, 0.5, \dots, 0.9,0.95\}$ 
    \end{tabular}
\end{center}




\begin{figure}[htb]
  \centering %
	\includegraphics[width=.9\linewidth, keepaspectratio]{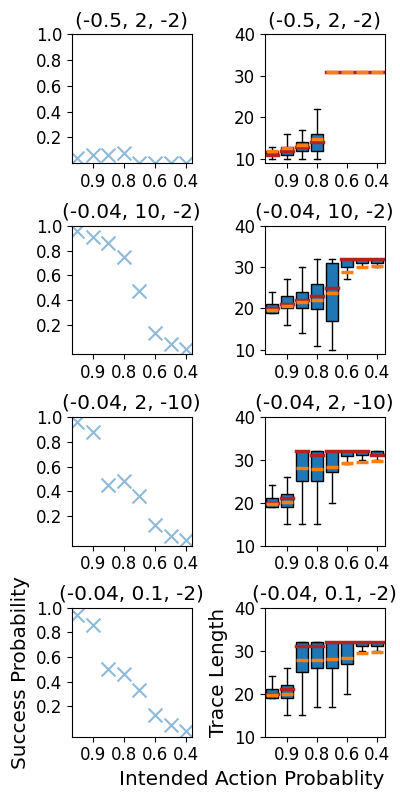}
    \caption{\small The left column shows the success probability and the right column shows the trace length with various reward structures and intended action probabilities $p_{\rm in}$. Higher negative rewards in intermediate states leads to lower mission accomplishment rate.}
    \label{fig:cheese_home_plot}
\end{figure}


Figure~\ref{fig:cheese_home_plot} shows the average success probability from a sample of simulations. Each row represents a distinct reward structure, and the x-axis of each sub-plot represents intended action probabilities $p_{\rm in}$. Note that when $r_{other}$ has large negative values, the success probability decreases as the $p_{\rm in}$ decreases, which is consistent with standard results in reinforcement learning. In contrast, the trace length decreases when $r_{other}$ has large negative values because the agent ends up in the fire. The lesson from this experiment is that poor planning in a difficult problem  overrides the benefit of the guarantee $\Psi^{C2H}$ is satisfied whenever the BT returns success.


\subsection{Action-Node Policies with BT Feedback}



The property that every successful trace of the BT satisfies the \LTL formula from which it was derived can be used to learn policies for the action nodes while the BT is running.  


\subsubsection{Algorithm}



Represent the policy for an action node using a probabilistic state-action mapping, $\pi:S\rightarrow \Delta(A)$, where $S$ is the set of possible states and $\Delta(A)$ is a probability distribution over actions. Thus, the policy is a conditional probability of action given the state, $\pi(s) = p(a|s)$ initialized with uniform action probability.

The learning algorithm lets the BT run a fixed number of episodes ($\xi$), where each episode ends when the BT returns success or failure. Recall that failure can occur if the time limit is reached. During each episode, store the trace as a sequence of time-index state-action pairs $[(s(0),a(0)),\dots,(s(m),a(m))]$ where $m$ is the trace length. At the end of each episode, update the state action table using
\begin{equation}
    p(a(t)|a(t)) \leftarrow \pi(a(t)| a(t)) + \mu^{m-t} * b,
    \label{eq1}
\end{equation}
where $\pi$ is the policy, (s(t), a(t)) is the state-action pair at time \textit{t} in the trace $\tau$, $\mu=0.9$ is the discount factor and $b$ is a binary variable which is translated to +1 (-1) when $\mathbb{T}_{\psi}$ returns success (failure). After the update, each $p(a|s)$ is renormalized so that $\sum_{a}p(a|s) = 1$. Since every successful trace is guaranteed to satisfy $\Psi^{\rm C2H}$, setting $b=1$ makes actions observed in the trace more likely. It is not true that every failed trace does not satisfy $\Psi^{\rm C2H}$, the setting $b=-1$ biases exploration to those policies that lead to success.

We can use the fact that $\Psi^{C2H}$ is composed of two PPATasks connected by the \textbf{U} operator to use feedback from the PPATask subtree to learn different policies for $\psi^{Cheese}$ and  $\psi^{Home}$. Since the two tasks are sequential, the $\psi^{Cheese}$ sub-tree needs to be successful for $\psi^{Home}$ to be learned. Let $p(a|s,C)$ represent the phase where the agent is seeking the cheese, and let $p(a|s,H)$ represent the phase where the agent has the cheese and is learning to return home.  The BT nodes ${\it Action}_{\rm Cheese}$ and ${\it Action}_{\rm Home}$ use $p(a|s,C)$ and $p(a|s,H)$, respectively. Divide the trace $\tau$ into two phases, $\tau = [\tau_{C},\tau_{H}]$, where the first part of the trace attempts to perform the Cheese task and the second part attempts to perform the Home task. 

Two conditions apply. First, if the cheese subtree never returns success, Eq.~(\ref{eq1}) updates $p(a|s,C)$ over the entire trace $\tau$. Second, if the cheese subtree returns success, the $\tau_C$ is the part of the trace up to when the task is successful and is used to update $p(a|s,C)$. The remaining part of the trace is $\tau_H$, which uses BT successes of failures to update $p(a|s,H)$.

\subsubsection{Experiment Design and Results}

\begin{figure}[htb]
  \centering %
	\includegraphics[width=.9\linewidth, keepaspectratio]{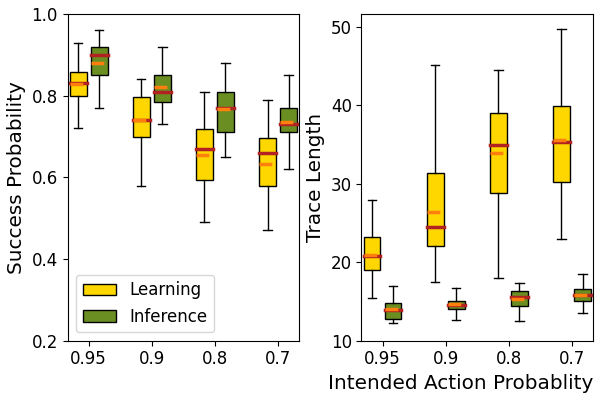}
    \caption{\small Success probabilities and trace lengths for learning phase and inference phase. }
    \label{fig:learning_results}
\end{figure}

All learning experiments use the following parameters, empirically selected to illustrate successful learning:  start a location $s_{(3,0)}$, $m=50$,$\xi=200$, and $\mu=0.9$ unless specified otherwise. Two dependent measures are used: \textit{learning success probability} and \textit{inference success probability}.  \textit{Learning success probability} is defined as the average success status of mission BT while the policies are being learned divided by the number of episodes. Similarly, the \textit{inference success probability} is the number of successful missions using learned policies, divided by the total number of simulation runs. The agent's starting location is randomized in the inference success evaluations, and 50 simulations were conducted using the learned policies.  

For the learning experiments, 50 independent simulations were conducted. Figure~\ref{fig:learning_results} compares the performance of policies during learning and inference settings. The golden boxes in the left sub-plot of Figure~\ref{fig:learning_results} show that learning $p(a|s)$ directly from the return status of the BT produces good policies for the sequential $\Psi^{C2H}$ problem. Similarly, the golden boxes in the right sub-lot of Figure~\ref{fig:learning_results} show that the \textit{trace length} increases with a decrease in intended action probabilities. Figure~\ref{fig:learning_results} depicts two distinct properties: a) the success probabilities are higher for the inference phase than the learning phase, and b) the trace length increases much more rapidly during the learning phase than the inference as the intended action probability decreases. These properties are seen because, during the inference phase, just one episode is sufficient to test the policy, whereas the learning phase requires many episodes where the uncertainty in the intended action probabilities accumulates with each episode. 
Comparing the results from Figure~\ref{fig:cheese_home_plot} to Figure~\ref{fig:learning_results} shows that using the direct feedback from the BT to create policies for the action nodes has higher success probabilities than using policy iteration when the rewards and goal do not align well.

\section{Fetch Robot Example}

This section demonstrates how a BT for an \LTL-based achievement goal works on the \textit{Fetch} robot.  A Fetch robot is a mobile robot with a manipulator arm that has a) 7 degrees of freedom, b) a modular gripper with easy gripper swapping, c) a torso with adjustable height, and d) an ability to reach items on the floor. The camera is at the head of the robot, so during manipulation tasks, its arm blocks its field of view. 


\subsection{Problem Specification}

The sequential problem for the robot is a variant of \textit{key-door}~\cite{minigrid} problem. A rectangular box of size 1ftx1ft is the \textit{active region} where its vision system is actively scanning, and the area outside the box is a \textit{passive region}. The task has three blocks with three different shapes: the red block is the key, the black block is the door, and the blue block is the prize. The goal for the robot is to a) locate the key and stack it on the top of the door block, b) move both key and door blocks together to the passive zone, and c) locate the prize and carry it to the passive zone. The robot perceives the world through its vision system and interacts with its arm. 

An \LTL formula for \textit{key-door} (KD) mission is
\[
\Psi^{KD} = \textbf{U} (\;\textbf{F} \; \psi^{Key})  (\;\textbf{U}\; \textbf{F} \; \psi^{Door}  \;\textbf{F} \; \psi^{Prize})
\]
where three PPA tasks have the structure from in Eq.~\ref{eq:PPATask}

\begin{eqnarray*}
\psi^{Key} &=& \vee (\wedge \it{NoErr} \; \it KeyStacked)(\wedge (\wedge \it NoErr \; IsKeyDoor) \\ && (\textbf{U} \; VisibleKeyDoor \; \wedge \; \textit{Action}_{KeyStacked} \; \it{NoErr})) \\ 
\psi^{Door} &=& \vee (\wedge \it{NoErr} \; \it KeyDoorPassive)(\wedge (\wedge \it NoErr \; KeyStacked) \\ && (\textbf{U} \; KeyStacked \; \wedge \; \textit{Action}_{KeyDoorPassive} \; \it{NoErr})) \\
\psi^{Prize} &=& \vee (\wedge \it{NoErr} \; \it PrizePassive)(\wedge (\wedge \it NoErr \; PrizeVisible) \\ && (\textbf{U} \; KeyDoorPassive \; \wedge \; \textit{Action}_{PrizePassive} \; \it{NoErr}))
\end{eqnarray*}

 
\noindent where the \textit{NoErr} proposition checks if the robot is throwing any system errors, \textit{KeyStacked} proposition checks if the key and door block are stacked together, \textit{IsKeyDoor} checks if the key and door block are on the table, \textit{VisibleKeyDoor} checks if the key and door are visible and not overlapping, \textit{KeyDoorPassive} checks if key and door are in passive area of the table, \textit{PrizePassive} checks if the prize block is in passive area, and \textit{PrizeVisible} checks if the prize block is in the active area. The state vector at time~$t$ is the vector of truth values for all the atomic proposition described above, $\mathbf{s}_t = [\mbox{\it NoErr}, \mbox{\it KeyStacked}, \dots, \mbox{\it PrizeVisible}]$. 

The action nodes $\textit{Action}_{KeyStacked}$, $\textit{Action}_{KeyDoorPassive},$ and $\textit{Action}_{PrizePassive}$ execute the plans to stack the key on top of the block, move the stack of key and door to the passive area, and move the prize to passive area, respectively. The sensing and plans were created using widely available Fetch robot libraries~\cite{wise2016fetch}.

\subsection{Experiment Design}
Two different modes of the experiment were conducted: \textit{baseline} and \textit{PPA-Task-LTLf} . For each mode, 25 independent robot trial was done. One trial corresponds to allowing the robot to perform the mission until it returns failure or success. In the \textit{baseline} mode, the robot tried executing the mission without using \LTL to BT decomposition, and the plans were connected using if-else code blocks. The second mode \textit{PPA-Task-LTLf} used the $\Psi^{KD}$ specification and the corresponding BT. For each mode, ten trials were completed without any external disturbances. For the other remaining 15 trials, a human physically interrupted the robot by removing or moving blocks. For each trial, the disturbance was only done once. The interruption was uniformly applied at each stage of the mission.   Recall that the \textbf{F}inally operator from the mission grammar can send a reset signal to its child sub-tree. In all experiments below, if some tasks fail the robot can retry once, which was chosen subjectively as the robot generally accomplishes the mission after one try. 


\subsection{Results}

The top part of Table~\ref{tab:missionBT} shows the performance of the robot for the key-door problem without using \LTL to BT decomposition. The robot failed to complete the mission when interrupted in the baseline condition no matter when the interruption occurred. In the \textit{baseline} mode experiments, the robot does not have the ability to resume from where it last failed because the baseline algorithm does not track details of its failures and attempt to correct constraint violations. Since the BT has a modular postcondition-precondition-action (PPA) structure, by design, it can resume from the previous failure point if allowed to retry. 

The bottom part of Table~\ref{tab:missionBT} shows the robot's performance when mission BT is used. Despite human interruptions at different stages of the mission, most of the time, the robot could complete the mission as it was allowed to do one retry. The robot failed once on the $\psi^{Door}$ task and twice on the $\psi^{Prize}$  task after interruptions due to constraint violations that could not be reversed.  The most important constraint violation was a violation of the \textit{NoErr} proposition, which encodes robot system errors that arise when the planning sub-system is unable to generate plans for the current system states.  The descriptive data in Table~\ref{tab:missionBT} suggest that the mission success rate is higher when the behavior tree implementation of the mission grammar was used, which was one of the reasons for implementing the mission grammar in a behavior tree. 

\begin{table}
\centering
\begin{tabular}{|l|c|c|l|l|}
 \hline
 \multicolumn{5}{|c|}{\textbf{Baseline Conditions}} \\\hline
 Status & Normal & \multicolumn{3}{|c|}{Human Disturbances In} \\[0.5ex]\hline
  \multicolumn{2}{|c|}{} & $Task^{\rm Key}$& $Task^{\rm Door}$ & $Task^{\rm Prize}$ \\
 \hline
   \textbf{Success} & 10 & 0 & 0 & 0 \\ 
 \hline
  \textbf{Failure} & 0 & 5 & 5 & 5 \\ 
  \hline\hline
  \multicolumn{5}{|c|}{\textbf{Behavior Tree Conditions}} \\\hline
 Status & Normal & \multicolumn{3}{|c|}{Human Disturbances In} \\[0.5ex]\hline
  \multicolumn{2}{|c|}{} & $Task^{\rm Key}$& $Task^{\rm Door}$ & $Task^{\rm Prize}$ \\
 \hline
   \textbf{Success} & 10 & 5 & 4 & 3 \\ 
 \hline
  \textbf{Failure} & 0 & 0 & 1 & 2 \\ 
  \hline
\end{tabular}
  \caption{Experiment results under a) the baseline conditions and b) the BT obtained from Mission and Task grammars.}
  \label{tab:missionBT}
\end{table}

\section{Summary and Future Work}
This paper presented a mission grammar designed for achievement-oriented goals that require temporal coordination among subtasks. The temporal constraints were formalized in the mission grammar, which produced linear-temporal logic formulas from a subset of LTL operators. The grammar for the tasks was constructed to use postcondition-precondition-action structures, allowing the construction of behavior trees that used these structures. Every successful trace produced by the behavior tree satifies the LTL goal.  

A key structure of the behavior trees is that the action nodes can create plans using off-the-shelf planners, which is in contrast to many previous work on converting LTL formulas into state machines. The examples presented were straightforward demonstrations for how existing planners could be used to implement the action nodes. Some properties of the resulting planners were demonstrated, specifically the risk of reward-goal misalignment if using MDP-based planners, the ability to use the return status of the behavior tree to train state-action policies, and the ability to retry subtasks to produce more resilient behaviors. The most important piece of future work is to encode sophisticated goals for real robots performing complicated tasks, and then identify state-of-the-art planners that are most compatible with the type of feedback provided by the behavior tree.


\begin{acks}
This work was supported by the U.S. Office of Naval Research (N00014-18-1-2831). The authors thank Elijah Pettitt, who was an undergraduate research assistant, for programming and running the experiments with the Fetch robot. 
\end{acks}



\bibliographystyle{ACM-Reference-Format} 
\bibliography{sample}


\end{document}